\documentclass[sigconf, nonacm]{acmart}
\setcopyright{none}
\AtBeginDocument{%
  }

\setcopyright{acmlicensed}
\copyrightyear{2018}
\acmYear{2018}
\acmDOI{XXXXXXX.XXXXXXX}
\acmConference[Conference acronym 'XX]{Make sure to enter the correct
  conference title from your rights confirmation email}{June 03--05,
  2018}{Woodstock, NY}
\acmISBN{978-1-4503-XXXX-X/2018/06}
\usepackage{booktabs}
\usepackage{multirow}
\usepackage{enumitem}
\usepackage{algorithm}
\usepackage{algpseudocode}

\begin{document}

\title{Jev-LiteMAS: Separating Decisions from Generation in Multi-Agent Workflows}

\author{Kaixiang Wang}
\affiliation{%
  \institution{Shanghai Jiao Tong University}
  \city{Shanghai}
  \country{China}
}

\author{Zhaojiacheng Zhou}
\affiliation{%
  \institution{Shanghai Jiao Tong University}
  \city{Shanghai}
  \country{China}
}

\author{Jiong Lou}
\affiliation{%
  \institution{Shanghai Jiao Tong University}
  \city{Shanghai}
  \country{China}
}

\author{Jie Li}
\affiliation{%
  \institution{Shanghai Jiao Tong University}
  \city{Shanghai}
  \country{China}
}
\renewcommand{\shortauthors}{Trovato et al.}

\begin{abstract}

Multi-agent LLM systems rely on generative models for
expert execution, workflow planning, and artifact review.
Yet many coordination operations require only bounded
decisions, such as selecting the next expert or judging
artifact quality. Profiling three representative MAS
frameworks, we find that routing and review account for
31.7--43.5\% of generative-LLM tokens. However, replacing
these operations with low-cost decision models introduces
two interface gaps: generative planners produce execution
plans and handoff instructions (\emph{planning gap}), while
reviewers provide actionable feedback for repairing flawed
artifacts (\emph{review gap})—outputs that bounded decisions
alone cannot provide.

We introduce Jev-LiteMAS, a multi-agent framework that
bridges both gaps by combining a lightweight decision model
(Jev) with generative execution and feedback. To address
the planning gap, Jev selects experts dynamically from the
task and prior expert outputs, using accumulated context to
ground subsequent execution without a pre-generated trajectory.
To address the review gap, Jev screens intermediate artifacts
against task criteria: passing artifacts proceed directly,
whereas flagged ones undergo LLM re-review for acceptance or
corrective feedback. Across four benchmarks and two backbones,
Jev-LiteMAS reduces generative-LLM tokens by 12.7--54.6\%
over LAS, maintaining accuracy within 1.34 percentage points
and lowering latency in seven of eight settings. Grounding
routing in substantive expert outputs prevents up to 18.0
points in accuracy loss.
\end{abstract}

\ccsdesc[500]{Do Not Use This Code~Generate the Correct Terms for Your Paper}
\ccsdesc[300]{Do Not Use This Code~Generate the Correct Terms for Your Paper}
\ccsdesc{Do Not Use This Code~Generate the Correct Terms for Your Paper}
\ccsdesc[100]{Do Not Use This Code~Generate the Correct Terms for Your Paper}

\keywords{LLM, Multi-agent, Decision model, jev}


\maketitle

\section{Introduction}
\label{sec:introduction}

LLM-based multi-agent systems (MAS) organize specialist agents through
role assignments, communication, and intermediate artifacts
~\cite{wu2024autogen,hong2024metagpt,li2023camel}.
Recent systems further optimize workflow construction and agent
communication~\cite{zhang2025aflow,zhang2025maas,zhang2025gdesigner}.
These workflows combine expert execution with repeated coordination.
Experts generate substantive content, such as code, API-driven
solutions, and research proposals. Coordination operations select
the next expert and assess whether intermediate artifacts can proceed.
Many of these operations produce only bounded outputs, such as an
expert identifier or a pass/escalate decision.
Using a generative LLM for every such decision adds model calls
between expert contributions. This difference in output requirements
suggests an opportunity to allocate model capabilities at the level
of workflow operations.

\begin{figure}[t]
    \centering
    \includegraphics[width=\columnwidth]
    {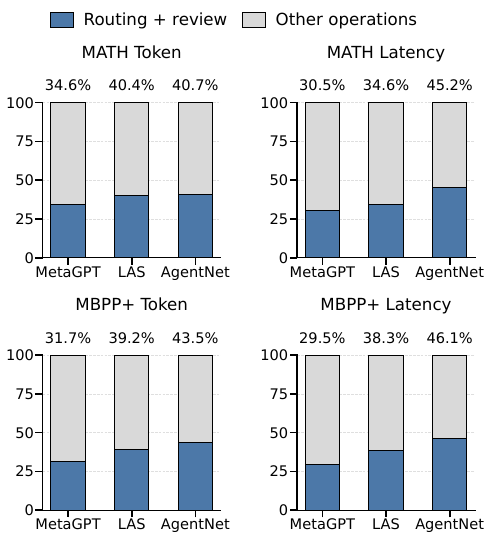}
    \caption{Routing and review account 
    of generative-LLM tokens and latency in our measured MAS configurations.
    Each bar is normalized by the total cost of its framework
    and across two benchmarks.}
    \label{fig:motivation_overhead}
\end{figure}

To quantify this opportunity, we profile three representative
MAS architectures, MetaGPT, AgentNet, and
LAS~\cite{hong2024metagpt,yang2025agentnet,xiang2026las},
using DeepSeek-V4-Flash as the shared generative backbone
on MATH~\cite{hendrycks2021math} and
MBPP+~\cite{austin2021program,liu2023evalplus}.
Figure~\ref{fig:motivation_overhead} shows that routing
and review account for 31.7--43.5\% of total generative-LLM
tokens across the six measured configurations.
Their recorded latency shares range from 29.5\% to 46.1\%.
Even in LAS, which dynamically adjusts workflow execution,
routing and review consume 40.4\% and 39.2\% of tokens
on MATH and MBPP+, respectively.
These results reveal substantial coordination overhead
across different MAS architectures and raise a more
specific question: which coordination operations require
open-ended generation, and which can be handled through
a lower-cost decision interface?

These observations motivate assigning routine routing and
review decisions to low-cost decision models, while retaining
generative models for substantive task execution.
Although prior studies have explored cost-aware model routing and
efficient MAS coordination~\cite{chen2024frugalgpt,
ong2025routellm,zhang2025agentprune,wang2025agentdropout},
integrating choice-based models into conventional multi-agent
workflows introduces two interface challenges.
\textbf{First, planning and expert handoffs.} In planner-driven
MAS, a generative planner produces a multi-step plan specifying task
decomposition, agent assignments, execution order, and handoff
instructions~\cite{hong2024metagpt,chen2024agentverse}.
In contrast, a choice-based decision model (e.g.,
Jev~\cite{typesafe2026jev}) only selects from predefined expert
candidates, leaving plan and instruction generation outside its
bounded interface.
\textbf{Second, review and corrective feedback.} Generative reviewers
assess intermediate artifacts and provide detailed revision guidance
when quality issues arise~\cite{madaan2023selfrefine,gou2024critic}.
Conversely, a choice-based judge like Jev only yields a bounded
verdict, unable to generate actionable strategies for subsequent repair.

These mismatches raise a central question:
\emph{how to reorganize multi-agent collaboration around bounded
decisions while preserving the planning context and generative feedback
required for effective execution and error recovery?}
This motivates our guiding principle:
\emph{allocate model capabilities according to each operation's
output requirements}. We organize collaboration around feedback-driven
decisions—using accumulated artifacts to ground execution—while
retaining generative capabilities for task execution and on-demand
corrective feedback.

We introduce Jev-LiteMAS, a multi-agent framework
that leverages the complementary strengths of generative
and decision models to reduce coordination overhead
while maintaining task quality. To address the planning gap, we develop
\emph{feedback-driven expert handoffs}.
Instead of generating a complete execution trajectory in
advance, Jev selects the next expert at each handoff
based on the overall task, previous expert outputs,
and the current execution state.
The selected expert receives the accumulated task context
and produces its contribution, allowing subsequent routing
decisions to adapt to actual execution progress.
To address the review gap, we introduce
\emph{Jev-gated LLM re-review}.
Jev first judges whether an intermediate artifact satisfies
the task-specific quality criteria, allowing qualified
artifacts to proceed without invoking a generative reviewer.
Artifacts that fail this judgment are escalated to an LLM
reviewer, which independently reassesses their quality and
either accepts them or generates actionable feedback for
revision by the responsible expert.
Revised artifacts are evaluated again before proceeding.
Together, these mechanisms replace routine generative
coordination with lightweight decisions while retaining
the contextual information and generative feedback
needed for effective collaboration and error recovery.

We evaluate Jev-LiteMAS on four benchmarks spanning mathematical
reasoning, code generation, tool use, and interdisciplinary idea
integration, using DeepSeek-V4-Flash and GPT-5 expert backbones.
Compared with LAS, Jev-LiteMAS reduces generative-LLM token usage
by 12.7--54.6\%, with accuracy differences within 1.34 percentage
points and lower end-to-end latency in seven of eight configurations.
Removing substantive expert feedback from routing decreases accuracy
by 18.00 points on IDRBench I3 and 9.02 points on
API-Bank~\cite{li2023apibank}, with nearly unchanged LLM call counts.
Under injected agent errors, Jev screening detects 91.33--95.33\%
of corrupted cases, while the cascaded configuration consumes
14.6--21.7\% fewer generative-LLM tokens than LLM judging.
These results connect the end-to-end efficiency gains to
feedback-aware coordination and selective generative review.

Our contributions are summarized as follows:
\begin{itemize}[leftmargin=2em]
    \item \textbf{Coordination-capability insight.}
    We quantify the generative overhead of routine MAS control
    and characterize the information required for bounded
    expert selection and actionable recovery.

    \item \textbf{Jev-LiteMAS architecture.}
    We develop an information-preserving workflow that seamlessly combines
    feedback-driven expert handoffs, bounded artifact screening,
    and conditional generative review and expert-guided repair.

    \item \textbf{Systematic evaluation.}
    We measure quality and resource use across four benchmarks
    and two backbones, and analyze workflow organization,
    semantic feedback, and review reliability through
    controlled ablations and error injection.
\end{itemize}

\section{Related Work}
\label{sec:related}

\noindent\textbf{LLM Multi-Agent Systems.}
LLM-based multi-agent systems coordinate specialized agents
through role assignments, intermediate artifacts, and shared
execution context. Recent research improves this coordination
at several levels. AFlow searches over code-represented
workflows, while MetaAgent constructs multi-agent systems
using finite-state machines~\cite{zhang2025aflow,zhang2025metaagent}.
MaAS learns a distribution of query-dependent architectures,
and G-Designer generates task-aware communication
topologies~\cite{zhang2025maas,zhang2025gdesigner}.
MasRouter jointly considers collaboration modes, agent
roles, and model selection~\cite{yue2025masrouter}.
AnyMAC predicts subsequent agents and selects context
from prior steps in sequential
collaboration~\cite{wang2025anymac}.
For execution efficiency, AgentDropout removes redundant
agents and communication, while LAS uses a lightweight
gate followed by LLM scheduling to adjust processing
depth~\cite{wang2025agentdropout,xiang2026las}.
AgentFlow coordinates planning, tool execution,
verification, and answer generation through evolving
memory, and ReAgent supports recovery by revisiting
earlier execution states~\cite{li2026agentflow,xinjie2025reagent}.
Together, these systems illustrate the growing role of
adaptive control and intermediate feedback in multi-agent
execution. Their varied coordination operations motivate
examining where bounded model decisions can be used
within an expert workflow.

\noindent\textbf{Jev and System-One Decision Models.}
Jev provides typed choices, rubric scores, and binary
judgments for software-controlled
decisions~\cite{typesafe2026jev}.
Independent evaluations examine its performance on
general decision benchmarks and text annotation,
including classification quality, inference cost,
and calibration~\cite{deusser2026jev,ibrahim2026decision}.
Further studies analyze decision stability under
changes to option semantics and assess typed decisions
across agent-harness tasks~\cite{sun2026typesafe,li2026fast}.
Beyond standalone judgments, Jev-Mem uses a decision
layer to manage memory operations~\cite{jiang2026jevmem}.
SeLMRoute extracts probabilistic semantic evidence
for model routing, and REFLEX explores bounded control
with generative fallback in
agents~\cite{perifanis2026selmroute,wu2026reflex}.
The contemporaneous S1-MAS framework likewise
allocates lightweight System-One decisions to
multi-agent coordination using an open decision
model~\cite{zhou2026s1mas}.
This literature provides a basis for studying typed
judgments in agent systems, where local
decisions affect downstream execution.

Our work Jev-LiteMAS integrates choice-based decisions into multi-agent workflows through feedback-driven expert handoffs and Jev-gated artifact review. It preserves generative capabilities for expert reasoning and corrective feedback, enabling selective LLM re-review while balancing task quality, review reliability, and execution efficiency.

\section{Problem Setting and Design Rationale}
\label{sec:problem}
This section formalizes the operation-level multi-agent workflow, examines the capability requirements of routing and review, and motivates our design.

\subsection{Operation-Level Multi-Agent Workflows}
\label{sec:problem:setting}

We consider a multi-agent workflow with a pool of
generative experts $\mathcal{E}=\{E_1,\ldots,E_n\}$.
Given a task $x$, the experts collaborate through
successive execution steps. At step $t$, the selected
expert $e_t\in\mathcal{E}$ produces an intermediate
artifact $y_t$ from the available execution state
$s_{t-1}$. The workflow then incorporates this
artifact into its updated state:
\begin{equation}
    y_t = G_{e_t}(s_{t-1}), \qquad
    s_t = U(s_{t-1},e_t,y_t),
    \label{eq:workflow}
\end{equation}
where $G_{e_t}$ denotes expert execution and $U$
denotes the workflow's state update. The observable
state contains the task, intermediate artifacts,
and relevant expert information.

We distinguish three operations within this
workflow: \emph{expert execution}, which generates
or revises task artifacts; \emph{expert routing},
which selects the next action; and \emph{local
review}, which evaluates an intermediate artifact.
At a routing point, the legal action set
$\mathcal{A}_t$ contains eligible experts and,
when permitted, a termination action. At a
review point, the reviewer determines whether
the artifact can proceed:
\begin{equation}
    \mathcal{A}_t \subseteq
    \mathcal{E}\cup\{\mathrm{STOP}\},\qquad
    d_t\in\{\mathrm{pass},\mathrm{escalate}\}.
    \label{eq:operations}
\end{equation}

These operations differ in their output requirements.
Expert execution requires open-ended generation,
while routing and initial review admit bounded
decision interfaces. This distinction creates an
opportunity to allocate model capabilities at the
level of workflow operations.

\subsection{Capability Mismatch in Workflow Control}
\label{sec:problem:mismatch}

\textbf{Planning and expert handoffs.}
Generative controllers can produce task-specific
plans and instructions for subsequent experts.
A choice-based model operates over a supplied
candidate set and returns a selected action:
\begin{equation}
    G_{\mathrm{plan}}(s_t)\in\Sigma^*,
    \qquad
    J_{\mathrm{route}}(s_t,\mathcal{A}_t)
    \in\mathcal{A}_t,
    \label{eq:routing_interface}
\end{equation}
where $\Sigma^*$ denotes the space of generated text sequences. This difference places greater importance on how the workflow provides legal actions and execution context. In particular, expert handoffs must remain informed by the latest artifacts, while the information needed for expert execution remains available to the next selected expert throughout subsequent execution.

\textbf{Review decisions and corrective feedback.}
Generative reviewers can assess an artifact and
provide task-specific feedback. A choice-based
reviewer produces a bounded decision:
\begin{equation}
    G_{\mathrm{review}}(s_t)=(v_t,c_t),
    \qquad
    J_{\mathrm{screen}}(s_t)=d_t,
    \label{eq:review_interface}
\end{equation}
where $v_t$ is the review verdict and $c_t$
denotes corrective feedback when needed.
The screening decision identifies artifacts
requiring further attention. Those artifacts
still need a generative reviewer to reassess
their content and provide actionable guidance
for subsequent repair.

Together, these two cases highlight the role
of information flow in capability allocation:
bounded decisions guide workflow execution,
while generative operations supply the content
needed to complete and revise tasks.

\subsection{Capability-Matched Workflow Design}
\label{sec:problem:rationale}

The preceding observations motivate an
operation-level division of responsibilities.
Generative experts retain task execution and
artifact revision. A choice-based model handles
bounded routing and initial review using the
observable execution state.

For coordination, the workflow selects subsequent
experts from the legal action set after observing
the latest execution results. This enables
successive handoffs to reflect the evolving
task state.

For review, the workflow adopts a conditional
generative path. The screening decision $d_t$
determines whether the artifact proceeds or
undergoes LLM re-review:
\begin{equation}
    (v_t,c_t)=
    \begin{cases}
        (\mathrm{pass},\varnothing),
        & d_t=\mathrm{pass},\\
        G_{\mathrm{review}}(s_t),
        & d_t=\mathrm{escalate}.
    \end{cases}
    \label{eq:cascaded_review}
\end{equation}
The generative reviewer reassesses escalated
artifacts and supplies corrective feedback
when needed. Subsequent revisions remain
the responsibility of the generative experts.

Follow these principles we present Jev-LiteMAS,
which integrates feedback-driven expert
handoffs with Jev-gated LLM re-review.
The following section presents the workflow
architecture and the detailed execution procedure.
\begin{figure*}[t]
    \centering
    \includegraphics[width=0.85\linewidth]
    {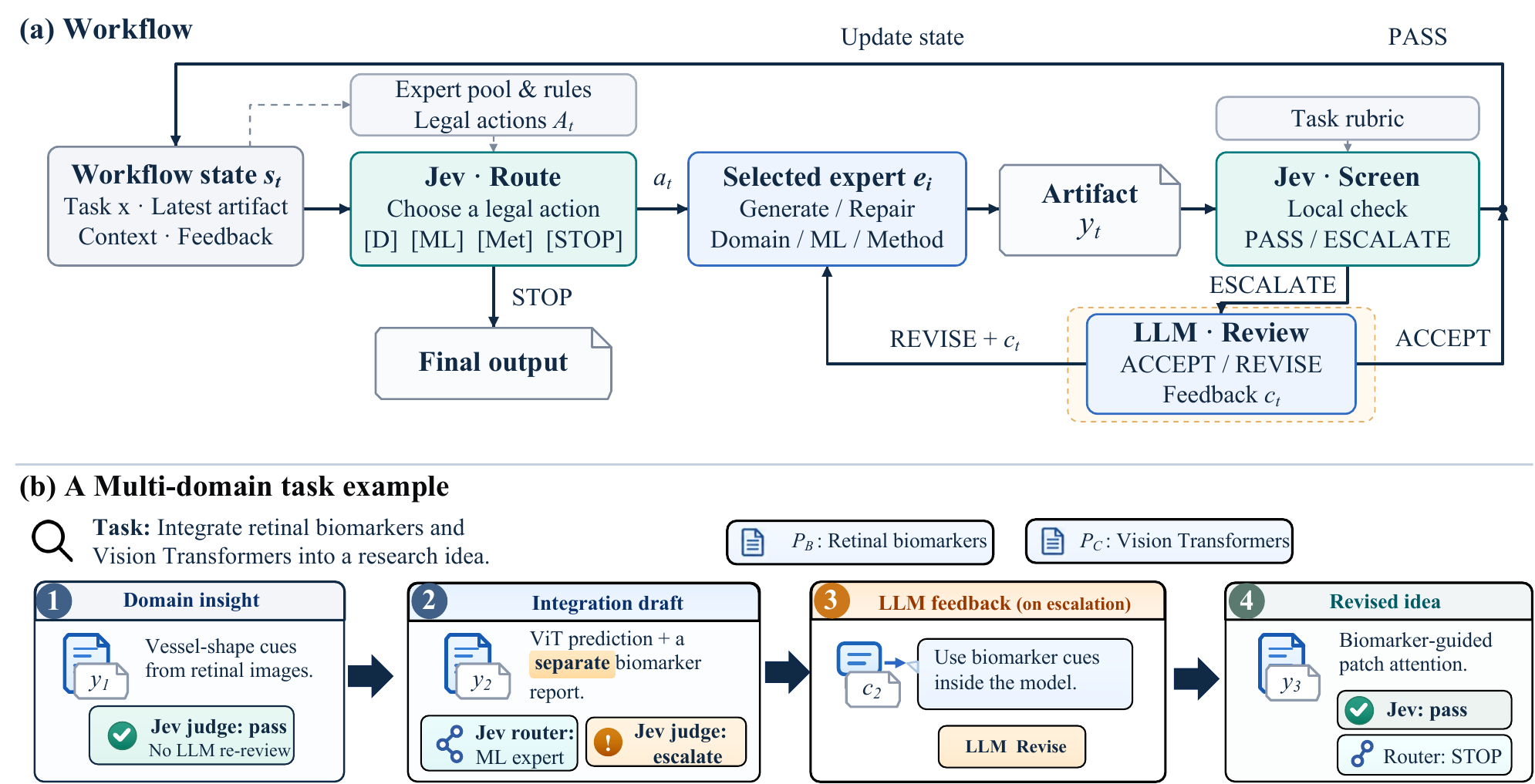}
    \caption{Overview of Jev-LiteMAS.}
    \label{fig:1}
\end{figure*}
\section{Jev-LiteMAS}
\label{sec:method}

This section presents Jev-LiteMAS, which coordinates generative
experts through feedback-driven handoffs and Jev-gated LLM re-review.
We first describe the shared execution state, then detail how routing,
screening, and corrective feedback shape the workflow.

\subsection{System Overview}
\label{sec:method:overview}

Figure~\ref{fig:1} illustrates the workflow. The system consists
of a pool of generative experts, two decision operations served by Jev,
and a generative reviewer for escalated artifacts. Experts create task
outputs and perform revisions. The routing operation chooses the next
expert from the legal candidates. The screening operation assesses
intermediate artifacts and decides whether they can proceed. When
screening requests further examination, the LLM reviewer evaluates the
artifact and provides feedback for subsequent execution.

Each expert $e\in\mathcal{E}$ has a configured generative backend $L_e$
and role instructions $p_e$. We write its execution interface as
\begin{equation}
    G_e(s)=\operatorname{Exec}(L_e,p_e,s),
    \label{eq:expert_configuration}
\end{equation}
where $s$ supplies the task, available artifacts, and any feedback
relevant to the current execution. Role instructions specify the
expert's responsibility, such as producing an artifact or revising
an existing one. The runtime manages model invocation and passes the
result back to the workflow. An expert's role remains associated
with its configuration across successive handoffs. The state
provides the changing part of its input: the user's request,
intermediate work, and feedback generated during execution.

The state also connects the two decision operations. The router
uses it to choose the next contributor, while the screener uses it
to assess the current artifact. Their prompts and output candidates
reflect these separate responsibilities. Both decisions can refer
to the same task and execution history, which makes the resulting
control actions consistent with the work already performed.

We use the state convention in Section~\ref{sec:problem}: expert $e_t$
produces $y_t$ from $s_{t-1}$, and $U$ incorporates the result into
$s_t$. The updated state is the common input to subsequent control
operations. It makes the latest artifact visible alongside the task
and prior execution context. At configured review points, screening
is resolved before the workflow proceeds with its next ordinary
handoff. Both Jev operations therefore act on artifacts generated
during the current execution, while generative experts retain the
context required to continue the task.

\subsection{Feedback-Driven Expert Handoffs}
\label{sec:method:handoffs}

\paragraph{Routing context and candidates.}
The next expert is selected after the workflow observes an execution
result. The runtime constructs the decision input from $s_t$ and the
legal set $\mathcal{A}_t$. The task describes the overall objective;
the current artifact records what has been produced; and the expert
descriptions indicate what each candidate can contribute. Candidate
identifiers connect the decision returned by Jev to executable expert
configurations. The set $\mathcal{A}_t$ follows the configured workflow
rules and contains eligible experts, with $\mathrm{STOP}$ included
when termination is an available action. Each candidate has an
identifier understood by the runtime and a description of its
assigned expertise. This correspondence is maintained across
routing steps, so the selected identifier can be used immediately
for dispatch. The candidate set determines the available workflow
transitions, while the current artifact helps Jev choose among them.

Jev chooses one action from these candidates:
\begin{equation}
    a_t=J_{\mathrm{route}}(s_t,\mathcal{A}_t),
    \qquad a_t\in\mathcal{A}_t.
    \label{eq:expert_selection}
\end{equation}
The routing request asks which expert should contribute next given
the task and available progress. As an expert completes its work,
the new artifact becomes evidence for later decisions. The router
can consequently select different experts as the task develops,
using the same bounded action interface at each decision point.
The decision may reflect the presence of an unfinished artifact,
a result requiring further inspection, or sufficient progress to
finish, depending on the candidates allowed by the workflow.
The process repeats after each completed expert execution, so new
intermediate results influence later decisions without a separate
generated plan for the entire trajectory.

\paragraph{Dispatch and execution context.}
If $a_t$ identifies an expert, the runtime resolves that identifier
to the corresponding model and role instructions and starts the next
execution step:
\begin{equation}
    e_{t+1}=a_t,\qquad
    y_{t+1}=G_{e_{t+1}}(s_t).
    \label{eq:expert_handoff}
\end{equation}
The selected expert receives the original task and the available
intermediate results through $s_t$. Its own role instructions establish
the expected contribution. This separation gives the controller a
compact decision interface while preserving task context for the
model responsible for generating content. The returned $y_{t+1}$ is
incorporated through $U$, making it available to the next routing or
review operation. The handoff carries the existing artifact forward;
the receiving expert can inspect and extend that work within its
configured responsibility. This is important when successive experts
address different aspects of the same task, since their outputs
remain connected through the shared state. It also preserves a
simple interface between the controller and the expert: the former
returns a legal action, and the latter generates the task content.

\paragraph{Successive handoffs.}
Handoffs continue as new artifacts change the state. For example,
a draft result may lead the router to select an expert responsible
for checking or refining it; the resulting feedback can then inform
another handoff. Such choices use the current trajectory and the
experts already configured in the workflow. If $a_t=\mathrm{STOP}$,
the runtime invokes the finalization procedure. A local review pass,
by comparison, only releases an artifact to the remaining workflow.
This distinction keeps artifact acceptance and task completion as
separate execution events. At a configured checkpoint, an artifact
must complete the local review path before ordinary coordination
resumes. Review-triggered revisions are therefore processed using
the feedback associated with that artifact, while the next normal
handoff operates on the updated state. The workflow can combine
several expert executions and local checks before finalization.

\begin{table*}[t]
\centering
\caption{End-to-end comparison across four benchmarks using DeepSeek-V4-Flash (top) and GPT-5 (bottom). ACC is in percent; Input, Output, and Total denote generative-LLM token counts; E2E is wall-clock latency in seconds; Calls denotes generative-LLM API calls per evaluation unit. Arrows indicate preferred directions.}
\label{tab:main_results}

\fontsize{8}{9}\selectfont
\setlength{\tabcolsep}{1.3pt}
\renewcommand{\arraystretch}{0.94}
\setlength{\aboverulesep}{0.24ex}
\setlength{\belowrulesep}{0.37ex}

\begin{tabular*}{\textwidth}{
@{\extracolsep{\fill}}l*{6}{r}
@{\hspace{5pt}\extracolsep{\fill}}*{6}{r}@{}
}
\toprule
\multicolumn{13}{@{}c@{}}{\bfseries DeepSeek-V4-Flash}\\
\midrule
Method & \multicolumn{6}{c}{MATH} & \multicolumn{6}{c}{MBPP+}\\
\cmidrule(lr){2-7}\cmidrule(l){8-13}
& ACC$\uparrow$ & Input & Output & Total$\downarrow$ & E2E$\downarrow$ & Calls$\downarrow$ & ACC$\uparrow$ & Input & Output & Total$\downarrow$ & E2E$\downarrow$ & Calls$\downarrow$\\
\midrule
Single-agent & 85.33 & 1234.8 & 944.7 & 2179.5 & 10.70 & 1.00 & 66.67 & 1257.6 & 632.8 & 1890.4 & 14.70 & 1.00\\
MetaGPT & 98.67 & 6754.1 & 3240.9 & 9995.0 & 36.07 & 4.00 & 82.67 & 7292.1 & 5774.8 & 13066.9 & 66.36 & 4.00\\
AFlow & 86.67 & 2248.9 & 1667.2 & 3916.1 & 17.09 & 1.00 & 73.33 & 2305.5 & 2906.9 & 5212.4 & 19.90 & 1.00\\
AgentNet & 96.67 & 3256.4 & 4467.6 & 7724.0 & 29.03 & 2.67 & 78.00 & 3610.7 & 4623.7 & 8234.4 & 43.91 & 3.13\\
LAS & 98.00 & 3792.0 & 2791.0 & 6583.0 & 26.15 & 3.03 & 84.67 & 5132.0 & 3587.0 & 8719.0 & 45.60 & 3.09\\
\textbf{Ours} & 97.33 & 3848.4 & 1584.8 & 5433.2 & 22.13 & 2.96 & 85.33 & 4136.0 & 2767.5 & 6903.5 & 38.67 & 2.96\\
\addlinespace[0.5pt]
\midrule
Method & \multicolumn{6}{c}{API-Bank} & \multicolumn{6}{c}{IDRBench I3}\\
\cmidrule(lr){2-7}\cmidrule(l){8-13}
& ACC$\uparrow$ & Input & Output & Total$\downarrow$ & E2E$\downarrow$ & Calls$\downarrow$ & ACC$\uparrow$ & Input & Output & Total$\downarrow$ & E2E$\downarrow$ & Calls$\downarrow$\\
\midrule
Single-agent & 35.28 & 4466.8 & 2543.9 & 7010.7 & 23.52 & 1.00 & 54.67 & 1849.0 & 1143.8 & 2992.8 & 14.73 & 1.00\\
MetaGPT & 62.47 & 21514.9 & 8579.4 & 30094.4 & 50.73 & 9.89 & 82.67 & 9342.1 & 7006.3 & 16348.4 & 86.61 & 4.00\\
AFlow & 45.62 & 4691.1 & 2553.8 & 7244.9 & 29.76 & 2.74 & 61.33 & 4257.6 & 3504.8 & 7762.3 & 43.79 & 2.00\\
AgentNet & 59.10 & 10206.3 & 5724.2 & 15930.6 & 38.25 & 6.38 & 77.33 & 14084.5 & 13313.2 & 27397.7 & 164.45 & 6.12\\
LAS & 66.52 & 11307.0 & 4986.0 & 16293.0 & 41.39 & 6.63 & 76.00 & 12700.0 & 9856.0 & 22556.0 & 133.52 & 5.56\\
\textbf{Ours} & 66.29 & 6552.8 & 3316.5 & 9869.3 & 35.34 & 4.00 & 75.33 & 4928.3 & 5305.5 & 10233.8 & 72.40 & 2.47\\

\specialrule{0.95pt}{1.3pt}{1.0pt}
\multicolumn{13}{@{}c@{}}{\bfseries GPT-5}\\
\midrule
Method & \multicolumn{6}{c}{MATH} & \multicolumn{6}{c}{MBPP+}\\
\cmidrule(lr){2-7}\cmidrule(l){8-13}
& ACC$\uparrow$ & Input & Output & Total$\downarrow$ & E2E$\downarrow$ & Calls$\downarrow$ & ACC$\uparrow$ & Input & Output & Total$\downarrow$ & E2E$\downarrow$ & Calls$\downarrow$\\
\midrule
Single-agent & 96.00 & 1129.0 & 1211.9 & 2340.9 & 13.24 & 1.00 & 74.00 & 1162.0 & 764.7 & 1926.7 & 9.32 & 1.00\\
MetaGPT & 100.00 & 6250.7 & 5062.7 & 11313.4 & 64.07 & 4.00 & 81.33 & 6804.0 & 4645.1 & 11449.1 & 55.05 & 4.00\\
AFlow & 95.33 & 2153.0 & 2195.3 & 4348.3 & 16.08 & 2.00 & 79.33 & 2191.1 & 2044.5 & 4235.6 & 24.47 & 2.00\\
AgentNet & 98.67 & 3211.8 & 2482.7 & 5694.4 & 27.14 & 2.52 & 82.00 & 3125.9 & 3468.8 & 7594.7 & 40.68 & 3.00\\
LAS & 98.00 & 4209.0 & 3454.0 & 7663.0 & 25.91 & 3.31 & 84.67 & 4842.0 & 3794.0 & 8636.0 & 43.60 & 3.18\\
\textbf{Ours} & 98.00 & 3572.0 & 2418.0 & 5990.0 & 30.04 & 2.98 & 83.33 & 4028.0 & 3514.0 & 7542.0 & 40.25 & 3.03\\
\addlinespace[0.5pt]
\midrule
Method & \multicolumn{6}{c}{API-Bank} & \multicolumn{6}{c}{IDRBench I3}\\
\cmidrule(lr){2-7}\cmidrule(l){8-13}
& ACC$\uparrow$ & Input & Output & Total$\downarrow$ & E2E$\downarrow$ & Calls$\downarrow$ & ACC$\uparrow$ & Input & Output & Total$\downarrow$ & E2E$\downarrow$ & Calls$\downarrow$\\
\midrule
Single-agent & 51.46 & 4644.9 & 2885.9 & 7530.8 & 30.92 & 1.00 & 46.67 & 1848.9 & 2025.7 & 3874.5 & 25.75 & 1.00\\
MetaGPT & 67.64 & 21882.7 & 11918.6 & 33801.3 & 152.63 & 10.90 & 62.00 & 10599.2 & 7540.7 & 18139.9 & 90.57 & 4.00\\
AFlow & 60.67 & 4636.4 & 2955.1 & 7591.5 & 36.83 & 2.83 & 51.33 & 6990.5 & 5339.8 & 12330.3 & 65.88 & 3.00\\
AgentNet & 67.19 & 13415.4 & 7576.8 & 20992.2 & 133.82 & 9.43 & 64.67 & 15388.9 & 9794.3 & 25183.2 & 127.38 & 5.97\\
LAS & 69.89 & 9637.0 & 6756.0 & 16393.0 & 67.39 & 5.90 & 63.33 & 14357.0 & 11083.0 & 25440.0 & 119.45 & 6.04\\
\textbf{Ours} & 68.99 & 5485.0 & 4589.0 & 10074.0 & 33.53 & 3.68 & 64.00 & 8951.0 & 6675.0 & 15626.0 & 75.57 & 3.37\\
\bottomrule
\end{tabular*}
\end{table*}

\subsection{Jev-Gated LLM Re-review}
\label{sec:method:review}

\paragraph{Artifact screening.}
At a designated review point, the screening request includes the
current task, artifact $y_t$, and the review criteria available in
$s_t$. Jev evaluates whether the artifact can proceed under these
criteria and returns a bounded decision:
\begin{equation}
    d_t=J_{\mathrm{screen}}(s_t)
    \in\{\mathrm{pass},\mathrm{escalate}\}.
    \label{eq:initial_screening}
\end{equation}
A passing artifact is released for continuation. The runtime records
$v_t=\mathrm{pass}$ and $c_t=\varnothing$ for this local decision.
An escalation routes the same task and artifact to the LLM reviewer.
The escalation decision initiates further examination; the LLM
then determines the review outcome for this branch. The screening
criteria follow the task setting being evaluated, such as whether
a solution satisfies its requirements or an output violates the
specified review conditions. The screening result is associated
with the current artifact and checkpoint. A later revision is
screened as a new artifact at its own review point.

\paragraph{Generative re-review.}
The generative reviewer examines the escalated artifact using the
available task context and review criteria. Its output contains a
verdict $v_t$ and feedback $c_t$:
\begin{equation}
    (v_t,c_t)=G_{\mathrm{review}}(s_t),
    \qquad d_t=\mathrm{escalate}.
    \label{eq:generative_rereview}
\end{equation}
The verdict decides whether the artifact may proceed or requires
revision. The feedback records the issues identified during review
and, when revision is needed, provides guidance for the next expert
execution. For instance, feedback on a program can identify an
unhandled input case, whereas feedback on a mathematical solution
can locate a questionable derivation step. This information is tied
to the submitted artifact so that the next model call can operate
on the existing work. The reviewer can also accept an escalated
artifact after reconsidering its content. In that case, the outcome
allows normal execution to resume, and no repair call is needed.
The verdict and feedback have separate uses: the former controls
the branch, and the latter informs the revision when requested.

\paragraph{Repair and re-entry.}
When the review outcome requests revision, the workflow assigns a
generative expert $e_{t+1}$ according to its repair protocol. We
denote the state with the new review information by $\widetilde{s}_t$:
\begin{equation}
    \widetilde{s}_t=s_t\oplus(v_t,c_t),\qquad
    y_{t+1}=G_{e_{t+1}}(\widetilde{s}_t).
    \label{eq:feedback_revision}
\end{equation}
Here $\oplus$ means that the runtime makes the review information
available to the assigned expert. The expert receives the task,
current artifact, and the returned feedback, and produces an updated
artifact. The runtime incorporates the revision through the state
update $U$ and applies the configured review procedure at the next
review point. The new artifact is treated as the result of another
expert execution, with its own state update and subsequent screening.
This preserves the link between the earlier feedback and the
revised content. If another problem is identified, the same
review--revision loop may continue within the configured limits.

This execution path also defines the scope of the screening gate.
A passing artifact proceeds without an LLM re-review at that
checkpoint. For escalated artifacts, the generative call supplies
an additional assessment and, when needed, a correction plan.
The resulting decisions and revisions remain visible in subsequent
workflow state, allowing ordinary expert coordination to resume.
This separation links the screening decision to the additional
content generated for the subsequent expert execution.

\subsection{Design Benefits}
\label{sec:method:benefits}

\textbf{Efficient coordination.}
Feedback-driven handoffs assign each decision to a bounded
action interface. Jev selects the next eligible expert using
the latest task state, while generative experts remain
responsible for producing and refining artifacts. This
division keeps routine coordination lightweight and allows
expert execution to use the accumulated task context.

\textbf{Selective generative review.}
In the common operating regime where most intermediate
artifacts satisfy the local review criteria, reviewing every
artifact with a generative LLM spends many calls on outputs
that already require no change. Jev screens these artifacts
first. Passing outputs proceed directly to the next workflow
step, and only escalated outputs invoke LLM re-review.
When acceptable artifacts dominate and the screening
false-escalation rate is low, the workflow avoids a
substantial fraction of generative review calls. The LLM
remains available to reassess escalated outputs and provide
corrective feedback when revision is needed.

The resulting workflow combines low-overhead coordination
with selective generative feedback.
Appendix~\ref{app:workflow_details} gives the full escalation
protocol and analyzes how artifact quality and screening
reliability affect review overhead.

\section{Experiment}
We systematically evaluate Jev-LiteMAS's task quality and execution cost through its separation of generative execution and bounded decision-making, and assess the contributions of individual components through ablation studies.
\subsection{Experimental Setup}
\label{sec:eval:setup}

\paragraph{Benchmarks and baselines.} Our evaluation covers four benchmarks across diverse domains:
MATH~\cite{hendrycks2021math} for mathematical reasoning,
MBPP+~\cite{austin2021program,liu2023evalplus} for code generation,
API-Bank~\cite{li2023apibank} for tool use, and
IDRBench I3~\cite{shen2025idrbench} for interdisciplinary
research idea integration.

We compare Jev-LiteMAS against five representative
baselines covering single-agent execution, multi-agent
collaboration, workflow optimization, and adaptive
scheduling. Single-Agent directly solves tasks with a
single LLM. MetaGPT~\cite{hong2024metagpt} coordinates
specialized agents through standardized workflows.
AFlow~\cite{zhang2025aflow} automatically searches and
optimizes agentic workflows.
AgentNet~\cite{yang2025agentnet} enables decentralized
agent collaboration through a structured network.
LAS~\cite{xiang2026las} dynamically schedules workflow
execution using a lightweight gate and an LLM scheduler.

\begin{table*}[t]
\centering
\caption{Ablations of workflow organization and routing
on IDRBench I3 and API-Bank. ACC is accuracy (\%);
Tok. counts generative-LLM tokens (K), excluding
decision-model tokens; E2E is latency (s), and Calls
counts generative-LLM API requests. SLM Router uses
Qwen3-1.7B-Instruct. \emph{Content-Blind} masks
substantive expert and tool outputs only in the router
input while preserving task information, legal actions,
execution metadata, and inter-expert communication.}
\label{tab:ablation}

\begin{minipage}[t]{0.49\textwidth}
\centering
{\bfseries (a) Workflow organization}\par\vspace{3pt}
{\fontsize{7.4}{9.0}\selectfont
\setlength{\tabcolsep}{1.1pt}
\renewcommand{\arraystretch}{1.12}
\begin{tabular*}{\linewidth}{
@{\extracolsep{\fill}}lrrrrrrrr@{}
}
\toprule
 & \multicolumn{4}{c}{IDRBench I3}
 & \multicolumn{4}{c}{API-Bank}\\
\cmidrule(lr){2-5}\cmidrule(l){6-9}
Method & ACC & Tok. & E2E & Calls
       & ACC & Tok. & E2E & Calls\\
\midrule
Static
 & 74.00 & 14.603 & 79.00 & 4.00
 & 61.80 & 15.178 & 35.71 & 4.00\\
Heuristic
 & 58.67 & 9.414 & 56.34 & 2.45
 & 64.27 & 7.564 & 31.53 & 3.49\\
Planner-Replan
 & 48.00 & 28.566 & 96.79 & 5.54
 & 53.93 & 21.873 & 47.65 & 8.31\\
\textbf{Jev-LiteMAS}
 & \textbf{75.33} & 10.233 & 72.40 & 2.47
 & \textbf{66.29} & 9.869 & 35.34 & 4.00\\
\bottomrule
\end{tabular*}
}
\end{minipage}%
\hfill%
\begin{minipage}[t]{0.49\textwidth}
\centering
{\bfseries (b) Routing and feedback}\par\vspace{3pt}
{\fontsize{7.4}{9.0}\selectfont
\setlength{\tabcolsep}{1.1pt}
\renewcommand{\arraystretch}{1.12}
\begin{tabular*}{\linewidth}{
@{\extracolsep{\fill}}lrrrrrrrr@{}
}
\toprule
 & \multicolumn{4}{c}{IDRBench I3}
 & \multicolumn{4}{c}{API-Bank}\\
\cmidrule(lr){2-5}\cmidrule(l){6-9}
Method & ACC & Tok. & E2E & Calls
       & ACC & Tok. & E2E & Calls\\
\midrule
LLM Router
 & 68.67 & 27.713 & 133.27 & 4.88
 & 65.62 & 15.418 & 56.13 & 7.56\\
SLM Router
 & 64.00 & 13.511 & 81.74 & 3.06
 & 61.35 & 11.764 & 39.15 & 4.81\\
Rule Router
 & 26.67 & 9.592 & 62.02 & 2.52
 & 38.43 & 8.762 & 35.75 & 3.87\\
Content-Blind
 & 57.33 & 17.437 & 98.23 & 2.46
 & 57.30 & 11.473 & 39.16 & 4.13\\
\textbf{Jev Router}
 & \textbf{75.33} & 10.233 & 72.40 & 2.47
 & \textbf{66.29} & 9.869 & 35.34 & 4.00\\
\bottomrule
\end{tabular*}
}
\end{minipage}
\end{table*}

\paragraph{Implementation and Metrics.}
We implement Jev-LiteMAS on top of the
AutoGen~\cite{wu2024autogen} framework, integrating
Jev-based expert routing and cascaded review into
multi-agent execution. In our main experiments, we use DeepSeek-V4-Flash~\cite{deepseek2026v4flash} and GPT-5~\cite{openai2025gpt5} as generative backbones, Jev~\cite{typesafe2026jev} as the lightweight decision model, and Qwen3-1.7B-Instruct as the SLM baseline. Subsequent analyses primarily use
the DeepSeek configuration to investigate the
contributions of workflow organization, routing
strategies, and selective review. Specifically,
we compare different MAS structures and routing
methods, inject errors into randomly selected
agents, and evaluate Jev and LLM judges under
clean and corrupted outputs. We further examine
sensitivity to generative-LLM call budgets,
heterogeneous expert configurations, and the
request-level overhead of GPT-5 and Jev.
We measure task accuracy (ACC), LLM
token usage (do not include SLM), end-to-end (E2E) latency, and the
number of generative-LLM API calls. For
error-injection experiments, we additionally
report attack success rate (ASR) and screening
false-positive and true-positive rates (FPR/TPR).
Detailed implementation settings, scoring rubrics, and evaluation
protocols are provided in
Appendix~\ref{app:experimental_protocol}.

\begin{table}[t]
\centering
\caption{Review robustness under injected agent errors on MATH and
MBPP+. ACC(\%) and ASR(\%); Tok. denotes generative-LLM
tokens; E2E is end-to-end latency (s). A dash denotes runs
without injected errors.}
\label{tab:review_robustness}
{\fontsize{8.2}{9.5}\selectfont
\setlength{\tabcolsep}{0.65pt}
\renewcommand{\arraystretch}{1.06}
\begin{tabular*}{\columnwidth}{
@{}l@{\extracolsep{\fill}}rrrr
@{\hspace{3pt}\extracolsep{\fill}}rrrr@{}
}
\toprule
 & \multicolumn{4}{c}{MATH}
 & \multicolumn{4}{c}{MBPP+}\\
\cmidrule(lr){2-5}\cmidrule(l){6-9}
Method
& ACC$\uparrow$ & Tok.$\downarrow$
& E2E$\downarrow$ & ASR$\downarrow$
& ACC$\uparrow$ & Tok.$\downarrow$
& E2E$\downarrow$ & ASR$\downarrow$\\
\midrule
Clean-MAS  & 96.67 & 6,752  & 26.77 & --
           & 86.00 & 7,794  & 45.92 & --\\
Attack-MAS & 46.67 & 6,389  & 24.62 & 50.00
           & 39.33 & 6,973  & 46.84 & 48.67\\
MAD        & 98.00 & 20,563 & 73.76 & 0.00
           & 85.33 & 28,705 & 98.76 & 1.33\\
G-Safe     & 69.33 & 6,884  & 31.33 & 27.34
           & 52.00 & 6,996  & 55.32 & 35.33\\
MAS-Shield & 91.33 & 9,746  & 40.51 & 5.33
           & 78.67 & 11,324 & 67.76 & 8.67\\
AgentMon.  & 92.67 & 13,569 & 53.45 & 4.00
           & 81.33 & 14,480 & 83.86 & 6.00\\
XG-Guard   & 80.00 & 7,081  & 25.84 & 16.67
           & 76.00 & 8,728  & 59.67 & 12.67\\
\addlinespace[1.1pt]
Clean-Jev  & 97.33 & 5,433  & 22.13 & --
           & 85.33 & 6,903  & 38.67 & --\\
Attack-Jev & 38.67 & 6,131  & 23.52 & 58.67
           & 42.00 & 6,542  & 34.93 & 44.67\\
Jev--LLM   & 94.67 & 10,138 & 45.14 & 2.67
           & 84.00 & 12,430 & 76.32 & 2.67\\
\addlinespace[1.1pt]
\textbf{Ours}
           & 94.00 & 8,657  & 34.19 & 3.33
           & 84.67 & 9,736  & 58.35 & 4.00\\
\bottomrule
\end{tabular*}
}
\end{table}

\subsection{End-to-End Quality and Efficiency}
\label{sec:eval:main}

Table~\ref{tab:main_results} compares Jev-LiteMAS with five baselines across four benchmarks and two generative backbones. Across tasks, Jev-LiteMAS maintains competitive accuracy with lower generative-LLM usage. Against LAS, it reduces total LLM tokens by 17.5--54.6\% with DeepSeek-V4-Flash and 12.7--38.6\% with GPT-5. The corresponding accuracy differences range from -0.67 to +0.66 percentage points on DeepSeek and from -1.34 to +0.67 on GPT-5. On DeepSeek, our method obtains the highest MBPP+ accuracy (85.33\%) and closely matches LAS on MATH, API-Bank, and IDRBench I3. On GPT-5, it matches LAS on MATH, improves IDRBench I3 accuracy by 0.67 points, and incurs modest reductions on MBPP+ and API-Bank. The resource savings also extend to execution time: Jev-LiteMAS reduces end-to-end latency relative to LAS on all four DeepSeek benchmarks and on three GPT-5 benchmarks. The largest reductions occur on IDRBench I3 with DeepSeek (45.8\%) and API-Bank with GPT-5 (50.2\%). GPT-5 MATH is the exception, with 30.04 s versus LAS's 25.91 s. MetaGPT achieves higher accuracy on DeepSeek IDRBench I3 (82.67\% versus 75.33\%) but uses substantially more tokens and API calls. These results demonstrate a favorable quality--efficiency trade-off through operation-level allocation of coordination and generation.

\subsection{Workflow and Routing Ablations}
\label{sec:eval:ablation}

Table~\ref{tab:ablation} examines the effects of workflow organization and routing decisions on IDRBench I3 and API-Bank. Among the four workflow organizations, Jev-LiteMAS achieves the highest accuracy on both tasks,
reaching 75.33\% and 66.29\%. Compared with Static, it improves accuracy by 1.33 and 4.49 points while reducing LLM tokens by 29.9\% and 35.0\%. Heuristic uses fewer tokens and runs faster, but its accuracy is lower,
particularly on I3 (58.67\%). Planner-Replan incurs the highest token usage and call counts, accompanied by the lowest accuracy in both tasks.

The routing ablation highlights the importance of
decision backends and execution feedback. Jev Router
outperforms LLM Router by 6.66 and 0.67 accuracy
points on IDRBench I3 and API-Bank, while reducing
generative-LLM tokens by 63.1\% and 36.0\%.
Compared with Qwen3-1.7B-Instruct (SLM Router),
Jev improves accuracy by 11.33 and 4.94 points,
with 24.3\% and 16.1\% fewer generative-LLM tokens
and lower end-to-end latency. Rule Router further
reduces token usage but substantially degrades accuracy.
Removing semantic feedback lowers accuracy from
75.33\% to 57.33\% on I3 and from 66.29\% to
57.30\% on API-Bank, with nearly unchanged LLM
call counts. These results highlight the capable bounded decisions and substantive
execution feedback for efficient routing.


\subsection{Jev-Review and Screening Performance}
\label{sec:eval:review}

\paragraph{Evaluation protocol and baselines.}
We evaluate local review on MATH and MBPP+ by injecting
task-relevant errors into an intermediate artifact from a
randomly selected expert, simulating hallucinations that
enter subsequent collaboration. All measurements use
DeepSeek-V4-Flash and report results at the task level.
Clean-MAS and Attack-MAS characterize the underlying MAS
before and after corruption; Clean-Jev and Attack-Jev
provide corresponding Jev-workflow references.
We compare five protective baselines:
Multi-agent Debate~\cite{du2024debate} performs multi-round
deliberation, G-Safeguard~\cite{wang2025gsafeguard} uses
topology-aware anomaly detection,
MAS-Shield~\cite{wang2025masshield} employs staged auditing,
AgentMonitor~\cite{chan2024agentmonitor} monitors execution
and supports corrections, and
XG-Guard~\cite{pan2026xgguard} detects anomalies through
bi-level graph representations.
We further include Jev--LLM Judge, which uses a
generative judge, and Jev-LiteMAS, which screens with
Jev and invokes LLM re-review on escalation.
We measure final accuracy (ACC), attack success rate (ASR),
generative-LLM tokens, and end-to-end latency; the separate
screening experiment reports FPR and TPR.

\paragraph{End-to-end comparison.}
Table~\ref{tab:review_robustness} summarizes task-level
performance under injected errors. Attack-MAS drops to
46.67\% and 39.33\% accuracy on MATH and MBPP+, with
attack success rates of 50.00\% and 48.67\%.
Jev-LiteMAS raises accuracy to 94.00\% and 84.67\%
and reduces attack success to 3.33\% and 4.00\%.
Against MAS-Shield, it gains 2.67 and 6.00 accuracy
points while using 11.2\% and 14.0\% fewer tokens.
Multi-agent Debate attains the highest accuracy and
lowest attack success rates among the protected
configurations, accompanied by substantially higher
resource use. Relative to Jev--LLM Judge, our method
maintains similar task accuracy and reduces latency
by 24.3\% and 23.5\%, respectively.


\begin{table}[t]
\centering
\caption{Jev, LLM, and SLM judging under clean and injected-error
conditions. ACC and Rate are percentages; Rate denotes FPR
($\downarrow$) for clean cases and TPR ($\uparrow$) for attacks.
Tok. counts generative-LLM tokens only; E2E is latency (s).
SLM uses Qwen3-1.7B-Instruct.}
\label{tab:screening_metrics}
{\fontsize{8.0}{9.2}\selectfont
\setlength{\tabcolsep}{1.6pt}
\renewcommand{\arraystretch}{1.06}
\begin{tabular*}{\columnwidth}{
@{}c@{\hspace{5pt}}l
@{\extracolsep{\fill}}rrrr@{}
}
\toprule
Case & Judge & ACC & Rate & Tok. & E2E\\
\midrule
\multirow{3}{*}{\shortstack[c]{MATH\\Clean}}
& Jev       & 97.33 & 4.00 & 5,433 & 22.13\\
& LLM Judge & 96.67 & 1.33 & 7,691 & 35.44\\
& SLM Judge & 98.00 & 0.00 & 5,135 & 23.32\\
\midrule
\multirow{3}{*}{\shortstack[c]{MATH\\Attack}}
& Jev       & 94.00 & 95.33 & 8,657 & 34.19\\
& LLM Judge & 94.67 & 96.67 & 10,138 & 45.14\\
& SLM Judge & 68.67 & 61.33 & 7,149 & 42.33\\
\midrule
\multirow{3}{*}{\shortstack[c]{MBPP+\\Clean}}
& Jev       & 85.33 & 6.67 & 6,903 & 38.67\\
& LLM Judge & 86.67 & 5.33 & 9,110 & 53.71\\
& SLM Judge & 84.67 & 0.67 & 7,678 & 41.17\\
\midrule
\multirow{3}{*}{\shortstack[c]{MBPP+\\Attack}}
& Jev       & 84.67 & 91.33 & 9,736 & 58.35\\
& LLM Judge & 84.00 & 92.00 & 12,430 & 76.32\\
& SLM Judge & 48.00 & 44.67 & 9,315 & 63.18\\
\bottomrule
\end{tabular*}
}
\end{table}

Table~\ref{tab:screening_metrics} compares Jev, LLM,
and SLM judging under clean and corrupted outputs.
Jev achieves true-positive rates of 95.33\% on MATH
and 91.33\% on MBPP+, approaching the LLM judge's
96.67\% and 92.00\%.
In contrast, the SLM Judge
achieves only 61.33\% and 44.67\% TPR, indicating
substantially weaker detection of injected errors.
Interestingly, SLM Judge has lower clean-case
false-positive rates (0.00\% and 0.67\%) than both
Jev (4.00\% and 6.67\%) and LLM Judge
(1.33\% and 5.33\%).
This contrast suggests that fewer false alarms do
not necessarily indicate reliable error detection.
Despite slightly higher false-positive rates than
LLM Judge, Jev reduces clean-case generative-LLM
tokens by 29.4\% and 24.2\%, with latency reductions
of 37.6\% and 28.0\%.
Under corruption, token savings remain 14.6\% and
21.7\%, while maintaining detection rates close
to those of the LLM judge.
These findings demonstrate a favorable balance between screening reliability and generative-review overhead, which is not consistently achieved by a SLM judge.


\begin{figure}[t]
    \centering
    \includegraphics[width=0.8\linewidth]
    {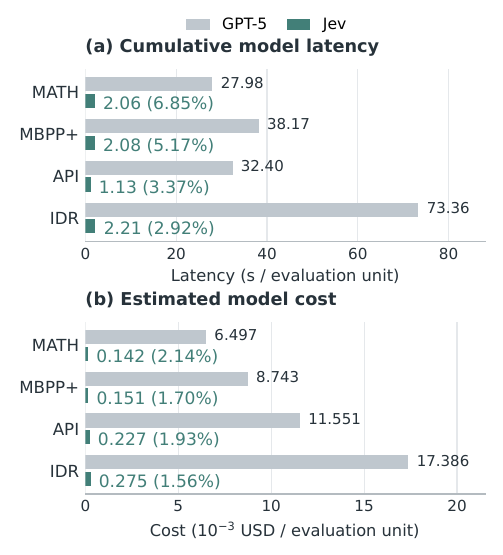}
    \caption{Latency and estimated model-cost breakdown
    of Jev-LiteMAS with GPT-5 experts and Jev routing
    across four benchmarks.
    (a) End-to-end latency and its GPT-5/Jev request-time
    components; (b) GPT-5 and Jev API fees.}
    \label{fig:decision_overhead}
\end{figure}
\subsection{Decision-Layer Overhead}
\label{sec:eval:overhead}

We analyze the end-to-end latency and estimated model cost of
Jev-LiteMAS with GPT-5 experts and Jev across four
benchmarks. Figure~\ref{fig:decision_overhead} separates the
contributions of generative GPT-5 requests and Jev decisions.
GPT-5 accounts for most of the execution time, while Jev adds only
1.13--2.21 seconds per evaluation unit. The Jev share of end-to-end
latency ranges from 2.92\% on IDRBench I3 to 6.85\% on MATH.
The monetary breakdown follows a similar pattern. Under the work fees, Jev accounts for 1.56--2.14\% of the
LLM model cost across the four benchmarks. Overall latency
ranges from 30.04 to 75.57 seconds, and the model cost from
\$0.00664 to \$0.01766 per evaluation unit. These measurements show
that the choice-based model layer introduces little direct overhead
compared with generative expert execution, supporting the efficiency
of operation-level capability allocation.

\subsection{Additional Analyses.}
We further examine the effect of generative-LLM call
limits and heterogeneous expert backbones on Jev-LiteMAS.
These experiments characterize quality--resource trends
under different execution budgets and model assignments
across expert handoffs; detailed results appear in
Appendix~\ref{app:extra_evaluations}.

\section{Conclusion}
\label{sec:conclusion}

We presented Jev-LiteMAS, a multi-agent framework
that leverages the complementary strengths of generative
and decision models to reduce coordination overhead.
Feedback-driven expert handoffs replace pre-generated
execution plans with dynamic routing informed by prior
expert outputs, while Jev-gated LLM re-review reserves
generative feedback for artifacts requiring further
examination or revision. Across four benchmarks and
two generative backbones, Jev-LiteMAS reduces generative-LLM
token usage by 12.7--54.6\% relative to LAS, with accuracy
differences within 1.34 percentage points and lower latency
in seven of eight configurations. Ablation studies further
highlight the importance of preserving substantive execution
feedback, demonstrating that lightweight decisions and
generative execution can jointly support efficient,
high-quality multi-agent collaboration.

\bibliographystyle{ACM-Reference-Format}
\bibliography{sample-base}
\clearpage

\appendix




\section{Limitations and Future Work}
\label{sec:limitations}

Jev-LiteMAS has two limitations.
First, feedback-driven routing operates over a predefined
pool of expert roles and legal actions. Although this
supports lightweight and structured coordination, the
framework does not dynamically create new experts or
reconfigure their roles for previously unseen task
requirements. Extending expert selection to more adaptive
collaboration structures is a promising direction.
Second, the efficiency of Jev-gated LLM re-review depends
on the distribution of intermediate artifact quality.
When a larger proportion of artifacts requires generative
review or revision, the achievable token and latency
savings may decrease. Future work could explore adaptive
review policies that account for task difficulty and
varying artifact quality.

\section{Additional Experimental Analyses}
\label{app:extra_evaluations}

We analyze the effects of generative-LLM call limits
and heterogeneous expert assignment on execution
quality and resource use.

\subsection{Sensitivity to Generative-LLM Call Budgets}
\label{app:call_budget}

We study how the maximum number of generative-LLM
calls affects execution quality and resource
consumption on IDRBench I3 and API-Bank.
Table~\ref{tab:appendix_budget_heterogeneous} reports
accuracy, token usage, latency, and actual LLM calls
under budgets of two, three, four, and six calls.

On I3, accuracy rises from 46.67\% at a budget of two
to 70.00\% at four, then reaches 73.33\% at six.
The corresponding average number of calls increases
from 2.00 to only 2.64, indicating that the workflow
does not exhaust the larger allowance on every task.
API-Bank follows a similar quality trend: accuracy
increases from 56.18\% to 64.72\% between budgets
two and four, and reaches 65.84\% at six. Its
average call count rises from 2.00 to 4.46.

Larger budgets also increase the opportunities for
further expert contributions, together with additional
token and latency costs. On API-Bank, token usage
and latency fluctuate across intermediate budgets,
reflecting changes in the executed trajectories.
Overall, the strongest quality gains occur when
moving beyond the tightest call limits.

\subsection{Heterogeneous Expert Backbones}
\label{app:heterogeneous}

We evaluate heterogeneous expert execution by
assigning two generative backbones (Deepseek V4-flash and GPT5), denoted A and B,
to successive expert steps. AAA keeps the same
backbone throughout the trajectory. ABB switches
from A to B at the first handoff and then continues
with B, while ABA switches backbone at every handoff.
All three configurations use the same Jev-based
coordination workflow.

As shown in Table~\ref{tab:appendix_budget_heterogeneous},
IDRBench I3 accuracy increases from 75.33\% with
AAA to 76.67\% with ABB and 78.00\% with ABA.
The higher I3 scores are accompanied by increased
token consumption, from 10.233K to 11.387K and
14.536K, respectively, and modestly longer
execution times.

On API-Bank, accuracy is 66.29\%, 66.07\%, and
64.49\% for AAA, ABB, and ABA. Token usage rises
from 9.869K to 12.924K and 13.518K, while average
calls increase from 4.00 to 4.23 and 4.39.

The results demonstrate that the workflow supports
switching expert backbones across handoffs. They
also reveal task-dependent trade-offs between
heterogeneous expert allocation, execution quality,
and model usage.

\begin{table}[t]
\centering
\caption{Sensitivity to generative-LLM call budgets and
heterogeneous expert backbones on IDRBench I3 and
API-Bank. ACC is in percent; Tok. denotes total
generative-LLM tokens in thousands; E2E denotes
end-to-end latency in seconds; Calls denotes average
generative-LLM API calls. In (b), AAA uses one backbone
throughout, ABB switches backbone at the first expert
handoff, and ABA switches at every handoff.}
\label{tab:appendix_budget_heterogeneous}
{\fontsize{7.1}{8.6}\selectfont
\setlength{\tabcolsep}{0.8pt}
\renewcommand{\arraystretch}{1.10}
\begin{tabular*}{\columnwidth}{
@{\extracolsep{\fill}}lrrrrrrrr@{}}
\toprule
& \multicolumn{4}{c}{IDRBench I3}
& \multicolumn{4}{c}{API-Bank}\\
\cmidrule(lr){2-5}\cmidrule(l){6-9}
Setting & ACC$\uparrow$ & Tok.$\downarrow$
& E2E$\downarrow$ & Calls$\downarrow$
& ACC$\uparrow$ & Tok.$\downarrow$
& E2E$\downarrow$ & Calls$\downarrow$\\
\midrule
\multicolumn{9}{@{}l}{
\textit{(a) Generative-LLM call budget}}\\
$B=2$ & 46.67 & 7.637 & 43.86 & 2.00
      & 56.18 & 10.713 & 16.12 & 2.00\\
$B=3$ & 62.67 & 9.496 & 56.01 & 2.26
      & 57.08 & 12.040 & 20.63 & 2.93\\
$B=4$ & 70.00 & 10.514 & 68.19 & 2.48
      & 64.72 & 14.553 & 18.34 & 3.96\\
$B=6$ & 73.33 & 11.298 & 71.88 & 2.64
      & 65.84 & 12.817 & 20.16 & 4.46\\
\midrule
\multicolumn{9}{@{}l}{
\textit{(b) Heterogeneous expert backbones}}\\
AAA & 75.33 & 10.233 & 72.40 & 2.47
    & 66.29 & 9.869 & 35.34 & 4.00\\
ABB & 76.67 & 11.387 & 75.97 & 2.49
    & 66.07 & 12.924 & 37.50 & 4.23\\
ABA & 78.00 & 14.536 & 78.62 & 2.52
    & 64.49 & 13.518 & 36.99 & 4.39\\
\bottomrule
\end{tabular*}
}
\end{table}

\section{Cascaded Review: Protocol and Analysis}
\label{app:workflow_details}
This section details the escalation and revision workflow, and analyzes how artifact quality and screening reliability jointly affect generative review overhead and error propagation.
\subsection{Escalation and Re-entry}

At a review checkpoint, Jev screens artifact $y_t$
using the context and review criteria in $s_t$.
A $\mathrm{pass}$ releases the artifact to ordinary
routing. An $\mathrm{escalate}$ forwards the same
information to $G_{\mathrm{review}}$, which returns
verdict $v_t$ and feedback $c_t$. If the LLM accepts
the artifact, routing resumes. If revision is
required, the runtime attaches $(v_t,c_t)$ to the
state and invokes the assigned expert, as formalized
in Eq.~\ref{eq:feedback_revision}. The revised
artifact enters the configured screening procedure
at its next checkpoint. Repeated review and repair
consume workflow budget. Local acceptance and the
$\mathrm{STOP}$ action remain separate: only the
latter enters task finalization.

\subsection{Escalation, Savings, and Reliability}

Let $\mathcal{R}$ be the completed screening
operations, including repeated checks after revision.
Every escalation triggers one LLM re-review, giving
the observed escalation fraction
\begin{equation}
\label{eq:escalation_fraction}
\widehat{\rho}_{\mathrm{esc}}=
\frac{\sum_{t\in\mathcal{R}}
\mathbf{1}\{d_t=\mathrm{escalate}\}}
{|\mathcal{R}|}.
\end{equation}

Let $\pi$ denote the fraction of artifacts requiring
correction, $r$ the detection rate
$\Pr(\mathrm{escalate}\mid\mathrm{error})$,
and $f$ the false-escalation rate
$\Pr(\mathrm{escalate}\mid\mathrm{acceptable})$.
The expected fraction of artifacts forwarded to
the LLM is
\begin{equation}
\label{eq:escalation_probability}
\rho_{\mathrm{esc}}=\pi r+(1-\pi)f.
\end{equation}
When most artifacts are acceptable ($\pi$ is small),
a low $f$ allows the majority to bypass generative
review. Thus, review-call savings follow directly
from the prevalence of acceptable outputs and the
screen's ability to release them.

For a simplified fixed per-call screening cost
$C_J$ and LLM review cost $C_L$, the expected
review-stage cost per checkpoint is
\begin{equation}
\label{eq:cascade_cost}
C_{\mathrm{cascade}}=C_J+\rho_{\mathrm{esc}}C_L.
\end{equation}
Relative to LLM review at every checkpoint, this
stage saves cost when
$C_J<(1-\rho_{\mathrm{esc}})C_L$.
Full-workflow accounting also includes routing,
expert generation, revisions, and repeated checks.

The main reliability risk is erroneous release.
Its unconditional probability is
\begin{equation}
\label{eq:missed_screening}
\Pr(\mathrm{error},\mathrm{pass})=\pi(1-r).
\end{equation}
These errors do not enter the LLM re-review branch.
Consequently, detection and false-escalation rates
must be assessed alongside final task quality and
resource use. The prevalence $\pi$ comes from
natural workflow traces; controlled error-injection
class ratios do not estimate it.

\subsection{Experimental Protocol}
\label{app:experimental_protocol}

\textbf{Dataset sampling.}
We evaluate all methods on identical, fixed subsets of four
benchmarks to ensure comparable experimental conditions.
For MATH~\cite{hendrycks2021math},
MBPP+~\cite{austin2021program,liu2023evalplus}, and
IDRBench I3~\cite{shen2025idrbench}, we use stratified random
sampling with seed 42, preserving the proportions of task
categories in each benchmark. We sample 150 instances from
each benchmark. For API-Bank~\cite{li2023apibank}, we sample
150 episodes using the same random seed, comprising 445
API-call prediction points in total. All compared methods
are evaluated on the same sampled instances.

\textbf{Task-specific evaluation.}
We evaluate the final task outputs using the native
evaluation criteria of each benchmark.
For MATH, we extract the final answer and compute accuracy
using mathematical equivalence checking, without requiring
exact textual matching.
For MBPP+, we execute generated programs using EvalPlus
and report pass@1, counting an instance as successful
only when it passes both the base and extended test suites.
For API-Bank, given the interaction history, the model
predicts the next API invocation. A prediction is counted
as correct only when the API name is correct and the
official checker validates the arguments and execution
result. We report strict accuracy over the 445 API-call
prediction points. For IDRBench I3, accuracy is the proportion of evaluation instances whose predicted integrable/non-integrable labels match the dataset ground truth, with invalid outputs counted as incorrect.

\textbf{Resource accounting.}
For each evaluation instance, we collect provider-reported
input and output token usage across the corresponding
generative-LLM requests and aggregate them at the task
level. We report the average input, output, and total
generative-LLM tokens per instance. Dollar-denominated
model costs are calculated from the recorded token usage
and the corresponding model pricing, then averaged over
the evaluation instances. End-to-end latency and the
number of generative-LLM API calls are recorded from
local execution logs for each complete task and averaged
over the sampled instances. For API-Bank, these resource
metrics are aggregated per episode rather than per
API-call prediction point. This accounting separates
final task accuracy from the computational resources
consumed by the complete multi-agent workflow.

\section{Workflow Pseudocode}
\label{app:workflow_pseudocode}

Algorithm~\ref{alg:jev_litemas} summarizes the complete Jev-LiteMAS workflow, including feedback-driven expert handoffs, Jev-gated LLM re-review, and corrective revision.

\begin{algorithm}[h]
\caption{Jev-LiteMAS Workflow}
\label{alg:jev_litemas}
\footnotesize
\begin{algorithmic}[1]
\Require Task $x$, expert pool $\mathcal{E}$, configuration $\Gamma$
\Ensure Final task output

\State $(s,e) \gets \Call{Initialize}{x,\Gamma}$

\While{\Call{BudgetAvailable}{s}}
    \State $y \gets G_e(s)$
    \State $s \gets U(s,e,y)$

    \Comment{Jev-gated LLM re-review}
    \If{\Call{ReviewPoint}{s}}
        \While{\Call{BudgetAvailable}{s}}
            \State $d \gets J_{\mathrm{screen}}(s)$
            \If{$d=\mathrm{pass}$}
                \State \textbf{break}
            \EndIf

            \State $(v,c)\gets G_{\mathrm{review}}(s)$
            \If{$v=\mathrm{accept}$}
                \State \textbf{break}
            \EndIf

            \State $e_r\gets\Call{AssignRepairExpert}{s}$
            \State $\tilde{s}\gets s\oplus(v,c)$
            \State $y\gets G_{e_r}(\tilde{s})$
            \State $s\gets U(\tilde{s},e_r,y)$
        \EndWhile
    \EndIf

    \If{\textbf{not} \Call{BudgetAvailable}{s}}
        \State \textbf{break}
    \EndIf

    \Comment{Feedback-driven expert handoff}
    \State $\mathcal{A}\gets\Call{LegalActions}{s}$
    \State $a\gets J_{\mathrm{route}}(s,\mathcal{A})$
    \If{$a=\mathrm{STOP}$}
        \State \Return \Call{Finalize}{s}
    \EndIf
    \State $e\gets a$
\EndWhile

\State \Return $\Call{HandleBudgetExhaustion}{s,\Gamma}$
\end{algorithmic}
\end{algorithm}

\end{document}